\pdfoutput=1

\documentclass[11pt]{article}

\usepackage{ACL2024}
\usepackage{times}
\usepackage{latexsym}

\usepackage[T1]{fontenc}

\usepackage[utf8]{inputenc}

\usepackage{microtype}

\usepackage{inconsolata}

\usepackage{amsmath}
\usepackage{amssymb}

\usepackage{graphicx}

\usepackage{algorithm}
\usepackage{algpseudocode}

\usepackage{multirow}
\usepackage{multicol}
\usepackage{tabularx}
\usepackage{mdframed}
\usepackage{svg}
\svgsetup{inkscapelatex=false}

\title{\ours: Symbolic Backward Chaining for Structured Natural Language Reasoning}

\author{Jinu Lee$^{1,2}$ \and Wonseok Hwang$^{1,3}$ \\
  $^{1}$ LBOX \quad  $^{2}$ University of Illinois Urbana-Champaign \quad $^{3}$ University of Seoul \\
  \texttt{\{jinulee.v, wonseok.hwang\}@lbox.kr}
}

\newcommand{\ours}{SymBa}
\newcommand{\ourslong}{Symbolic Backward Chaining}

\newcommand{\stdev}[1]{\tiny$\pm$#1\small}

\begin{document}
\maketitle
\begin{abstract}
To improve the performance and explainability of LLM-based natural language reasoning, \textit{structured reasoning} can be applied to generate explicitly structured proofs. Among different methods for structured reasoning, we specifically focus on backward chaining, where the proof goal is recursively decomposed to subgoals by searching and applying rules. We argue that current LLM-based backward chaining systems (\textit{e.g.} Least-to-most prompting and LAMBADA) are \textit{incomplete}, as they omit crucial algorithmic components identified from the classic backward chaining algorithm in computational logic (SLD Resolution). To this end, we propose a novel backward chaining system, \textbf{\ours} \ (\ourslong), which integrates a symbolic solver and an LLM. In \ours, the solver controls the proof process, and the LLM is only called when the solver requires new information to complete the proof. Empowered by completeness, \ours\ achieves a significant improvement in seven deductive, relational, and arithmetic reasoning benchmarks compared to the baselines.\footnote{We publicly disclose our code, data, and prompts for reproduction in the following \href{https://github.com/lbox-kr/symba}{repository}.}
\end{abstract}

\section{Introduction}


Large language models (LLMs) trained with massive amounts of natural language text have shown remarkable reasoning ability in various fields, including logical and arithmetic reasoning \citep{DBLP:journals/tmlr/WeiTBRZBYBZMCHVLDF22, DBLP:conf/nips/KojimaGRMI22}. However, autoregressively generated explanations as in Chain-of-thoughts might contain factual and logical errors, which tend to be more covert as LLMs scale up \citep{zhou2024larger}.

To enhance the accuracy and explainability of natural language reasoning, \textit{structured reasoning} has been frequently explored as an alternative. In this task, one must provide an explicitly structured explanation, \textit{i.e.} a \textit{proof tree} (also known as \textit{entailment tree}). These structured explanations offer high interpretability by showing how premises connect to intermediate and final conclusions \citep{dalvi-etal-2021-explaining, hong-etal-2022-metgen}.

Among popular approaches for structured reasoning, we focus on \textit{backward chaining} \citep{DBLP:books/daglib/0024693}. Backward chaining reasoners start from the goal and apply rules that decompose the goal into a set of subgoals. It is known to be efficient as it does not require a combinatorial search to generate the next step  \citep{kazemi-etal-2023-lambada}. Consequently, previous works have proposed LLM-based backward chaining \textit{systems}, which utilize few-shot LLMs to execute subtasks of the backward chaining process \citep{kazemi-etal-2023-lambada, DBLP:conf/iclr/ZhouSHWS0SCBLC23, khot2023decomposed}.



However, we argue that popular LLM-based backward chaining systems, namely Least-to-most prompting \citep{DBLP:conf/iclr/ZhouSHWS0SCBLC23} and LAMBADA \citep{kazemi-etal-2023-lambada}, are \textit{incomplete}. We compare their implementation to a classic backward chaining algorithm from computational logic---SLD Resolution \citep{kowalski1974predicate}---and provide minimal examples that show their incompleteness in Section \ref{sec:baselines}.

To address this issue, we propose \textbf{\ours} (\ourslong), a method that applies an SLD resolution-based symbolic solver directly to natural language reasoning. In \ours, the solver controls the proof process, and the LLM is only called when the solver requires new information to complete the proof. By this novel solver-LLM integration, \ours\ benefits from both the completeness of the SLD resolution and the natural language reasoning capability of LLMs.

\ours\ outperforms baselines on answer accuracy, proof accuracy, and efficiency in seven benchmarks from deductive, relational, and arithmetic reasoning. Empirical results show that Least-to-most prompting suffers from low proof accuracy in complex problems. LAMBADA, on the other hand, cannot handle relational and arithmetic reasoning properly. We claim that these are the direct consequences of their incomplete design. 

In summary, our contributions are as follows.
\begin{itemize}
    \item We inspect the incompleteness of previous LLM-based backward chaining systems (Least-to-most and LAMBADA) by comparing its algorithmic components to SLD resolution.
    \item We propose \textbf{\ours}, an LLM-based backward chaining system controlled by a symbolic solver.
    \item We show that \ours \ outperforms the baselines in various reasoning tasks by leveraging the completeness of the solver.
\end{itemize}

\section{Background}

\textbf{SLD Resolution} \citep{kowalski1974predicate} is the backward chaining algorithm for \textbf{logic programs}.

\subsection{Logic programming}

Logic programming is a programming paradigm for computing formal logic \citep{wielemaker:2011:tplp, lifschitz2019answer}. In logic programming, each \textit{rule} defines a logical implication relation between predicate \textit{terms}. The \textit{implied} term on the left-hand side is the \textit{head}, and the \textit{condition} terms on the right-hand side are referred to as \textit{subgoals}. \textit{Fact} is a special type of rule with no subgoals, meaning that the head term is unconditionally true. For instance, Rule 1 in Figure \ref{fig:sld_resolution} denotes that if there exists an \texttt{X} that is young and round (\textit{subgoals hold}), then Charlie is cold (\textit{head implied}).


\subsection{SLD Resolution algorithm}
\label{sec:backgroundsolver}

\begin{figure}[!htbp]
    \centering
    \includegraphics[width=\columnwidth]{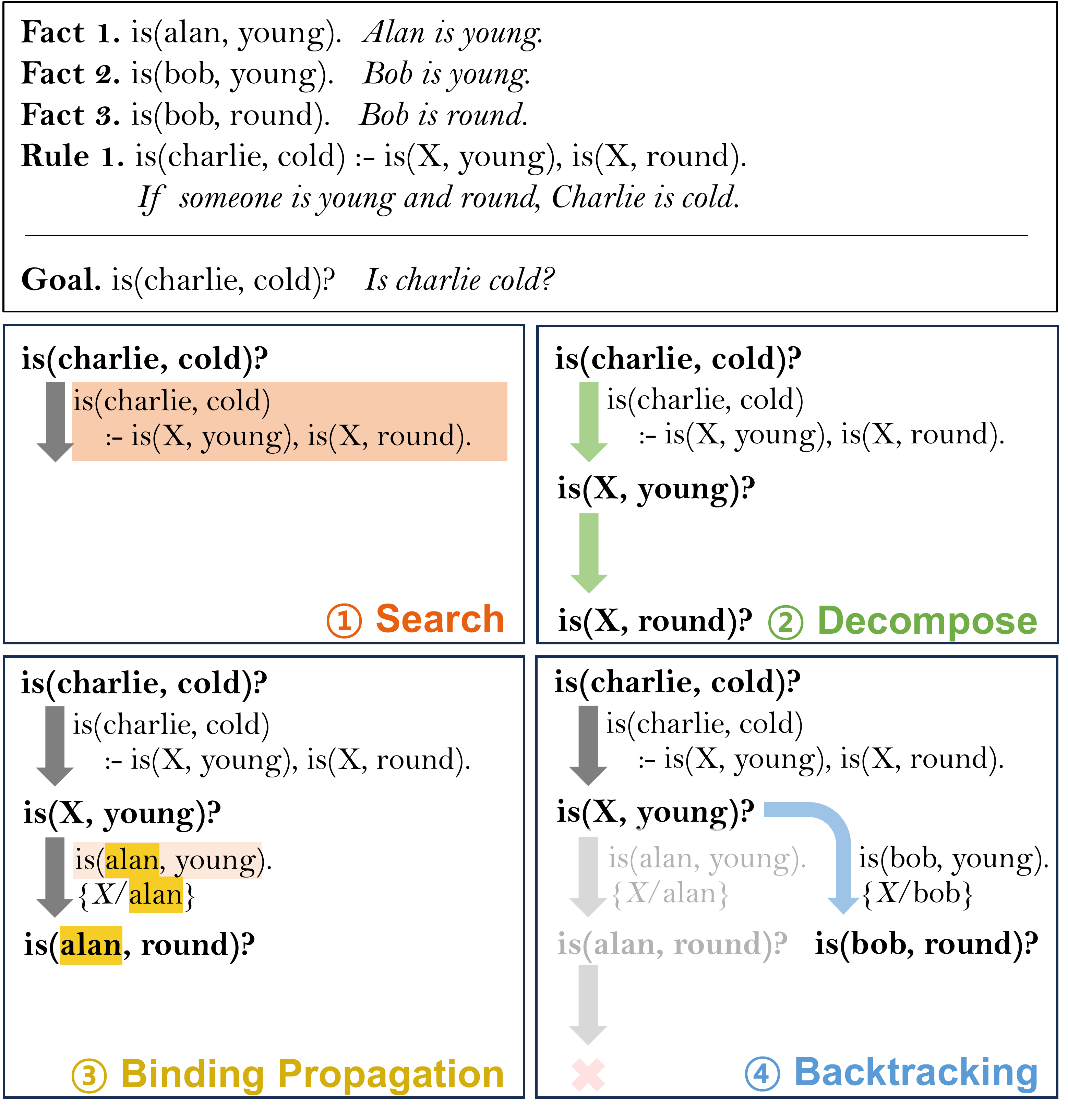}
    \caption{Example of a ProofWriter \citep{tafjord-etal-2021-proofwriter}-style problem written in both logic program and natural language (\textit{italic}). The four main steps of SLD Resolution, \textit{Search}, \textit{Decompose}, \textit{Binding Propagation} (between subgoals), and \textit{Backtracking}, are shown using this example.}
    \label{fig:sld_resolution}
\end{figure}

SLD Resolution algorithm recursively searches the valid proof for the \textit{goal} term using given rules. It can be viewed as a depth-first search algorithm with four key steps, \textit{Search}, \textit{Decompose}, \textit{Binding propagation}, and \textit{Backtracking}.

\textbf{Search} The proof process begins by searching for rules and facts that could support the goal. This is done by checking if there is a substitution of variables (\textit{binding}) that makes the goal and the rule head identical, \textit{i.e.} if the goal and the rule \textit{unifies}.

\textbf{Decompose} Once a unifying rule is found, the goal is broken down into the rule’s subgoals. These subgoals are added to the stack, and the proof is complete when all these subgoals are either proven or refuted.

\textbf{Binding propagation} Both goals and rules may contain variables. When a variable’s binding (antecedent) is determined during the proof, it must be propagated to other instances of the same variable to satisfy the coreferential constraints. In SLD resolution, binding propagation happens in three directions, from goal to subgoal, between subgoals, or subgoal to goal.

\textbf{Backtracking} If there are no rules that can prove the goal, the proof fails. In this case, the prover must backtrack and attempt alternative decompositions and bindings until a valid proof is found.

Consider the example in Figure \ref{fig:sld_resolution}. For the \textit{Search} step, the only rule that unifies to the given goal $\texttt{is}(\texttt{charlie}, \texttt{cold})$ is Rule 1. When we \textit{decompose} Rule 1, we get two subgoals $\texttt{is}(X, \texttt{young})$ and $\texttt{is}(X, \texttt{round})$. Initially, the first subgoal can be proved by binding $X/\texttt{alan}$, which is then \textit{propagated} and updating the second subgoal to $\texttt{is}(\texttt{alan}, \texttt{round})$. However, as this bound goal fails, \textit{backtracking} is required to explore other possible bindings for the first subgoal such as $X/\texttt{bob}$, which will eventually prove the goal.


Appendix \ref{sec:appendix-solver} presents a formal description of the algorithm.


\begin{figure*}[t]
    \centering
    \includegraphics[width=1.0\textwidth]{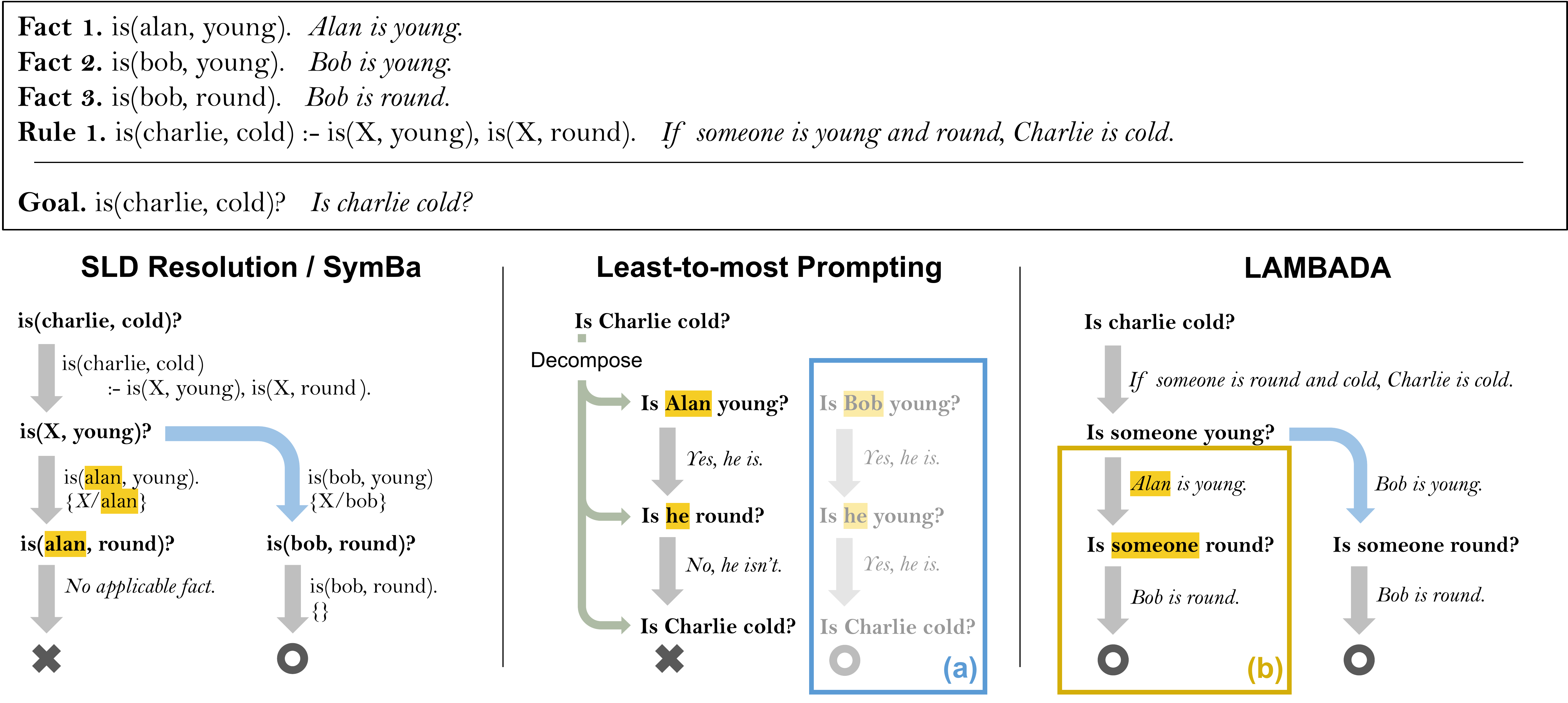}
    \caption{Comparison between SLD Resolution (and \ours), Least-to-most, and LAMBADA. Bindings $\{X/\texttt{alan}\}$ and $\{X/\texttt{bob}\}$ both apply to the first subgoal of Rule 1, but $\{X/\texttt{alan}\}$ fails to prove the second subgoal. While SLD Resolution and \ours\ traverse both possibilities and reach the correct conclusion with the correct proof, \textbf{(a)} lack of backtracking in Least-to-most might discard the correct trajectory, and \textbf{(b)} lack of binding propagation in LAMBADA might lead to an inaccurate reasoning step.}
    \label{fig:baselines}
\end{figure*}

\section{Methods}

\begin{figure*}[t]
    \centering
    \includegraphics[width=\linewidth]{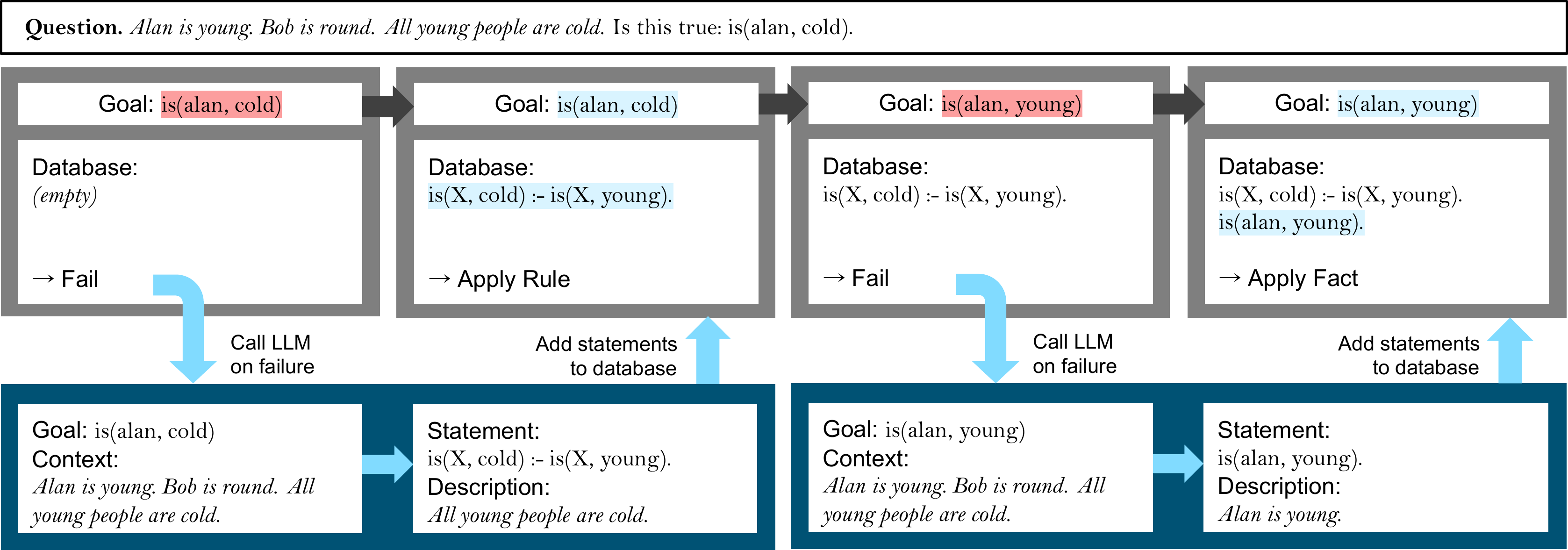}
    \caption{Overview of \ours. In \ours, a symbolic SLD Resolution solver (gray) controls the proof process. When a goal is not provable by the solver alone, an LLM (navy) is instructed to generate a single reasoning step which is then added to the symbolic solver's database (working memory).}
    \label{fig:overview}
\end{figure*}

\subsection{Baselines}
\label{sec:baselines}

We analyze two popular natural language-based backward chaining methods as our baseline, namely \textbf{Least-to-most prompting} \citep{DBLP:conf/iclr/ZhouSHWS0SCBLC23} and \textbf{LAMBADA} \citep{kazemi-etal-2023-lambada}.

\subsubsection{Least-to-most prompting}

\textbf{Least-to-most prompting} is a two-stage task decomposition method, consisting \textit{Decompose} and \textit{Solution} stage. In the initial \textit{Decompose} stage, the LLM is instructed to decompose the given question into subquestions and order them from least complicated to most. The subquestions are passed to the \textit{Solution} stage, where they are answered conditioned on both the problem and previous subquestion-answer pairs.

\textit{Decompose} and \textit{Solution} stages of Least-to-most prompting directly correspond to Decompose and Search steps of SLD resolution, respectively. Also, as the subquestions are answered by conditioning on the previous answers, it can be seen as implicitly performing binding propagation using the LLM's inherent coreference resolution ability.

The incompleteness of Least-to-most prompting comes from the fact that it does not allow \textit{backtracking} even if the decomposition is inaccurate. Figure \ref{fig:baselines}(a) depicts a scenario where two possible bindings exist for a subgoal but one eventually fails. In this case, Least-to-most cannot correct its decomposition even if it has failed to find a valid proof. As accurate decomposition is challenging when the reasoning path is long or when multiple plausible paths exist \citep{patel-etal-2022-question, DBLP:conf/iclr/Saparov023}, we show Least-to-most's proof accuracy is significantly harmed due to the failure in the Decompose stage (Section \ref{sec:proof-accuracy}). 

\subsubsection{LAMBADA}

\textbf{LAMBADA} implements a modular backward chaining approach that operates on pure natural language. When given a goal, it tests all facts and rules against the goal to find one that applies (\textit{Selection}). If a matching fact is retrieved, it stops recursion (\textit{Fact Check}). Instead, if a matching rule is retrieved, they are decomposed into subgoals (\textit{Decompose}). When multiple rules apply to the current goal, LAMBADA backtracks to traverse all possible reasoning trajectories.

While LAMBADA overcomes the limitation of Least-to-most prompting by implementing backtracking, LAMBADA fails to address binding propagation properly as it only implements the binding propagation from goal to subgoals. As a result, LAMBADA is inherently incapable of various types of reasoning including relational reasoning that requires binding between \textit{bridging entities} of subgoals (Figure \ref{fig:l2m-shortcut}) and arithmetic reasoning that requires binding propagation from subgoal to goal to pass the intermediate results up the tree (Figure \ref{fig:lambada}). Indeed, in the original paper, LAMBADA was only tested with deductive reasoning benchmarks \textit{without} bridging entities or arithmetic reasoning.

Besides the binding propagation problem, LAMBADA does not implement disjunction.\footnote{\textbf{Limitations} section, bullet 3 of \citet{kazemi-etal-2023-lambada}.} As a result, the behavior when the rule and goal have different signs is \textit{undefined}, as such cases require transforming conjunctive ($\land$) rules into disjunctive ($\lor$) ones by De Morgan's laws.

\subsection{Proposed method}
\label{sec:ours-method}

\subsubsection{\ourslong}


To overcome the limitations described above, we propose \textbf{\ours} (\ourslong), which directly integrates an SLD Resolution solver and an LLM for backward chaining in a coroutine (Figure \ref{fig:overview}).

Initially, the solver cannot prove the provided goal because its symbolic \textit{database} (working memory) is empty. To make progress, the solver calls the LLM to check if there is a rule or a fact in the natural language descriptions that might unify with the failed goal. When the LLM generates a unifying statement, the solver retries proving the failed goal with the new statement. The process is continued until the topmost goal is proved, or every possible reasoning path fails.
\begin{figure}[t]
    \centering
    \includegraphics[width=0.95\columnwidth]{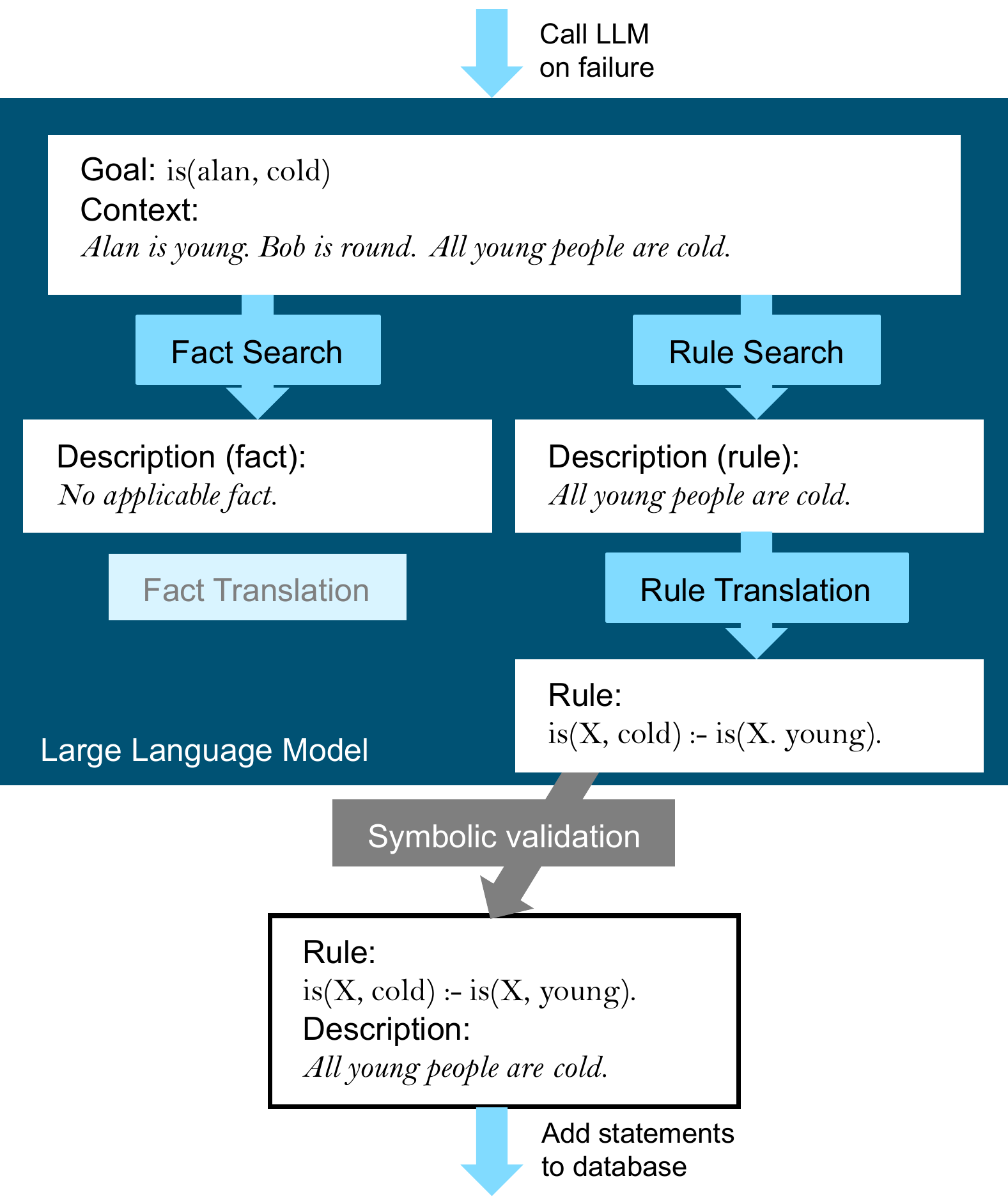}
    \caption{Brief illustration of the modules in \ours's single statement generation procedure. Search modules retrieve plausible reasoning steps from the context, which are translated into symbolic form by translation modules. Statements that pass the Symbolic Validation module are added to the solver's database.}
    \label{fig:modules}
\end{figure}

Delegating the proof control to a solver has numerous advantages. Most importantly, these solvers are sound and complete, guaranteeing correct explanations, provided that the symbolic statements are accurate. Furthermore, solver operations are lightweight compared to computationally intense LLM inferences.



\subsubsection{Single-step statement generation}
\label{sec:modules}

In \ours, the LLM is instructed to generate a logic program statement that can prove the current subgoal. Similarly to previous work on structured reasoning that adopts a modular strategy \citep{DBLP:conf/iclr/CreswellSH23, kazemi-etal-2023-lambada}, we divide the single-step statement generation process into five modules. Fact/Rule Search, Fact/Rule Translation, and Symbolic Validation (Figure \ref{fig:modules}).

\textbf{Fact/Rule Search} In the first stage, the LLM is prompted with the symbolic goal term and the natural language description of facts and rules, and retrieves ones that might prove the goal.

\textbf{Fact/Rule Translation} Subsequently, the LLM is given the goal and the natural language rule (obtained from the Search module) and generates a symbolic statement.

\textbf{Symbolic Validation} As a final step, \ours\ checks the translated facts and rules if they are (1) syntactically correct and (2) unify with the goal, which ensures that the translated statements can \textit{prove} the goal term. Note that this step is purely symbolic and does not require any LLM inference.

\section{Experimental settings}
\subsection{Benchmarks}
\label{sec:benchmarks}


\textbf{Deductive reasoning} Four representative benchmarks for deductive reasoning, namely the ProofWriter family (ProofWriter, Birds-Electricity, ParaRules) \citep{tafjord-etal-2021-proofwriter, ijcai2020p537} and PrOntoQA \citep{DBLP:conf/iclr/Saparov023}, are tested. Each instance is formulated as a binary classification task, deciding whether the given query can be proved according to the given rules and facts or not (\textit{closed-world assumption}).


\textbf{Relational reasoning} CLUTRR \citep{sinha-etal-2019-clutrr} is a relational reasoning benchmark based on human-written stories about family relations.  For our experiments, we reformulate the task into true/false form, where two entities and a relation are presented and one should predict if the given relation can be deduced from the story.

\textbf{Arithmetic reasoning} We use two popular arithmetic benchmarks, namely MAWPS \citep{koncel-kedziorski-etal-2016-mawps} and GSM8k \citep{cobbe2021gsm8k}. For both benchmarks, the goal is to predict the correct numeric answer for a short question.

For all benchmarks, performance is evaluated based on \textit{task accuracy}, which measures whether the predicted answer matches the gold label (true/false for deductive or relational tasks, and numerical for arithmetic tasks). Additionally, we manually assess \textit{proof accuracy} by verifying that every step in the proof is both correct and relevant \citep{DBLP:conf/iclr/Saparov023, kazemi-etal-2023-lambada}.

More information, including data statistics, few-shot example construction, logic program representation, and evaluation methods, can be found in Appendix \ref{sec:appendix_dataset}.

\subsection{Solver}

To implement the algorithm described in Section \ref{sec:backgroundsolver}, we develop an SLD Resolution-based solver in Python with necessary extensions, such as negation handling and arithmetic operations.


\subsection{Single-step statement generation}

To reproduce baselines and implement \ours, we use three open- and closed-sourced state-of-the-art LLMs: GPT-4 Turbo \citep{achiam2023gpt}, Claude 3 Sonnet \citep{anthropic_claude_2023}, and LLaMa 3 70B Instruct \citep{adams2024llama}.


For each module of \ours, few-shot demonstrations were sampled from each training split and manually converted to logic programs. To increase robustness, we adjust the few-shot examples to suppress hallucinations. Details can be found in Appendix \ref{sec:negative-examples}.

\section{Results}
\label{sec:result}
\begin{table*}[ht]
    \small
    \centering
    \begin{tabular}{l|l|c|c|c|c|c|c|c}
        \multirow{2}{*}{Model} & \multirow{2}{*}{Method} & \multicolumn{4}{c|}{Deductive} & Relational & \multicolumn{2}{c}{Arithmetic}\\
        \cline{3-9}
          &  & ProofWriter & BirdsElec & ParaRules & PrOntoQA & CLUTRR & MAWPS & GSM8k \\
          \hline \hline
          
         \multirow{3}{*}{GPT-4}
         & Least-to-most & 71.5& 88.2& 71.8& 87.5& 81.5& 84.3& 60.6\\
         & LAMBADA & 69.7& 83.4& 59.7& 96.0& 73.8 & 0.0 & 0.0 \\
         & \ours & \textbf{79.8}& \textbf{94.4}& \textbf{79.2}& \textbf{96.3}& \textbf{84.3}& \textbf{86.7} & \textbf{63.8}\\
        \hline \hline
        
         \multirow{3}{*}{Claude-3}
         & Least-to-most & 60.3 & 75.7 & 54.0 & 86.0 & 77.0 & \textbf{94.2} & 59.3 \\
         & LAMBADA & 69.3& 62.7 & 57.7 & 67.0 & 69.0 & 0.0 & 0.0 \\
         & \ours & \textbf{77.6} & \textbf{77.3} & \textbf{69.0} & \textbf{91.0} & \textbf{85.0} & \textbf{94.1} & \textbf{67.4} \\
        \hline \hline
        
         \multirow{3}{*}{LLaMa-3}
         & Least-to-most & 61.4 & 71.0 & 66.7 & \textbf{95.0} & 72.0 & \textbf{89.0} & 61.5 \\
         & LAMBADA & 64.0 & 82.3 & 62.1 & 90.8 & 73.3 & 0.0 & 0.0 \\
         & \ours & \textbf{70.4} & \textbf{92.9}& \textbf{71.7}& 93.3 & \textbf{90.5} & 87.9 & \textbf{67.0} \\
    \end{tabular}
    \caption{Average answer accuracy (\%) on four runs per each benchmark, LLM model, and reasoning method. Boldface indicates that the accuracy is significantly higher than others (confidence 95\%). LAMBADA predicts nothing in arithmetic benchmarks, resulting in zero accuracy. Complete results are shown in Appendix \ref{sec:appendix_fullresult}.}
    \label{tab:main}
\end{table*}


\subsection{Answer accuracy}
\label{sec:task-performance}

The main results are presented in Table \ref{tab:main}.
Among the three backward chaining systems compared (Least-to-most prompting, LAMBADA, and \ours), \ours \ demonstrates the strongest performance in diverse types of reasoning (deductive, relational, and arithmetic) and with different LLMs.



\begin{figure}[t]
    \centering
    \includegraphics[width=\columnwidth]{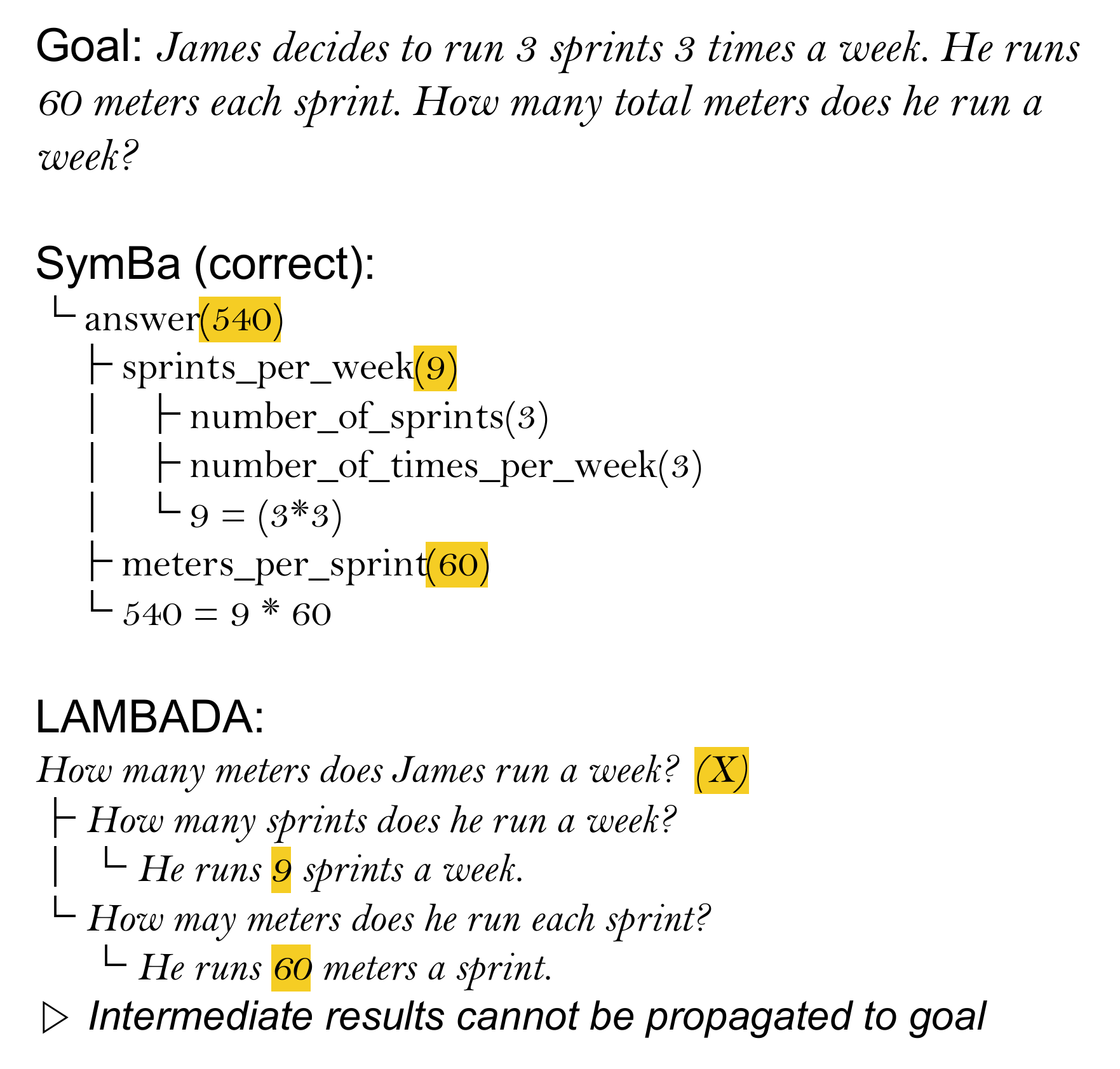}
    \caption{Example of LAMBADA's failure in GSM8k. While it can derive correct intermediate values, the lack of binding propagation from subgoal to goal will disallow them to be combined in higher nodes.}
    \label{fig:lambada}
\end{figure}

As LAMBADA does not implement binding propagation, LAMBADA cannot answer any arithmetic reasoning questions (Figure \ref{fig:lambada}). For CLUTRR, LAMBADA achieves higher answer accuracy than the random baseline (50.0), but it is only superficial because LAMBADA cannot apply coreferential constraints (further discussed in Section \ref{sec:proof-accuracy}).

The performance of \ours\ and baselines in ProofWriter is further analyzed in Table \ref{tab:deductive-negation}. We divide ProofWriter questions into \texttt{$\exists$Proof} questions that have a valid proof that either proves or disproves the goal, and \texttt{$\nexists$Proof} questions that cannot be proved or disproved due to lack of relevant information. \texttt{$\exists$Proof} questions are again separated into \texttt{$\exists$Negation} if the proof includes at least one negation (\textit{not}) and \texttt{$\nexists$Negation} otherwise. For example, the question in Figure \ref{fig:baselines} is in both \texttt{$\exists$Proof} and \texttt{$\nexists$Negation} because there is a valid proof that proves the goal, which does not contain any negation.

Least-to-most achieves low accuracy in the \texttt{$\nexists$Proof} set, \textit{i.e.} it frequently outputs a proof to an unprovable goal. On the other hand, \ours\ and LAMBADA achieve near-perfect scores in \texttt{$\nexists$Proof}, indicating that multi-depth decomposition and backtracking enhance the precision of the generated explanations. Although Least-to-most's accuracy seems to be high in the \texttt{$\exists$Proof} set, we show that the generated explanations are often incorrect, shadowing the accuracy gain (Section \ref{sec:proof-accuracy}).

\begin{table}[t]
    \small
    \centering
    \begin{tabular}{l|c|c|c|c}
          \multirow{2}{*}{Method} & \multicolumn{2}{c|}{\texttt{$\exists$Proof}} & \multirow{2}{*}{\texttt{$\nexists$Proof}} & \multirow{2}{*}{Overall} \\
          & \texttt{$\exists$Neg.} & \texttt{$\nexists$Neg.}  &  & \\
          \hline \hline
        \# examples & 97 & 76 & 127 & 300 \\
          \hline
         Least-to-most & 77.6 &  72.4 & 65.6 & 71.5\\
          LAMBADA & 4.7 & 73.2 & 98.2 & 69.7 \\
          \ours & 72.2 & 59.6 & 97.4 & 79.8
    \end{tabular}
    \caption{Fine-grained answer accuracy (\%) for ProofWriter (All systems use GPT-4 Turbo). Least-to-most demonstrates significantly low performance in \texttt{$\nexists$Proof} set, and LAMBADA suffers handling negation (\texttt{$\exists$Negation}).}
    \label{tab:deductive-negation}
\end{table}

As mentioned in Section \ref{sec:baselines}, LAMBADA cannot properly handle cases where the goal and the rule's sign disagree. The result shows that LAMBADA's accuracy significantly drops in \texttt{$\exists$Negation}, which explains the performance gap between \ours\ and LAMBADA in deductive benchmarks without binding propagation. 

\subsection{Proof accuracy}
\label{sec:proof-accuracy}

One of the key benefits of structured reasoning is that it generates more inspectable outputs \citep{ribeiro2023street}. In this section, we analyze the \textit{proof accuracy} of three backward chaining systems in four benchmarks. Following \citet{kazemi-etal-2023-lambada}, 30 proofs with correct answers are sampled from \texttt{$\nexists$Neg} ($\subseteq$ \texttt{$\exists$Proof}) and examined to see if they include any false intermediate statements or exclude necessary reasoning steps.

\begin{figure}[!t]
    \centering
    \includegraphics[width=\columnwidth]{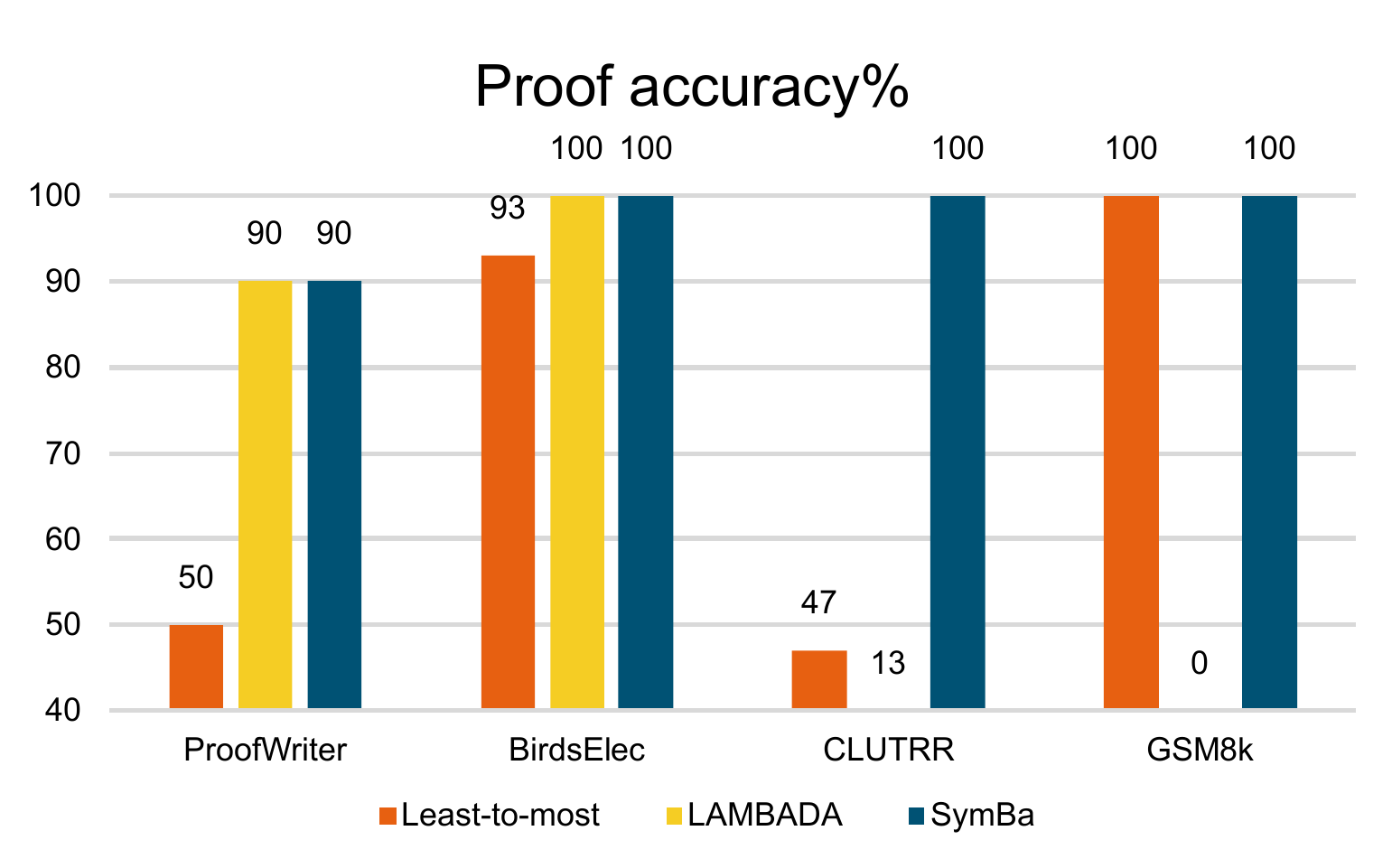}
    \caption{Proof accuracy on four reasoning benchmarks, using GPT-4 Turbo. Least-to-most achieves low proof accuracy in all benchmarks, while LAMBADA suffers in relational and arithmetic reasoning tasks.}
    \label{fig:proofaccuracy}
\end{figure}

Results are presented in Figure \ref{fig:proofaccuracy}. It is shown that \ours\ generates the most accurate proofs, where Least-to-most and LAMBADA prompting demonstrates significantly degraded proof accuracy in specific tasks.

\begin{figure}[tb]
    \centering
    \includegraphics[width=\columnwidth]{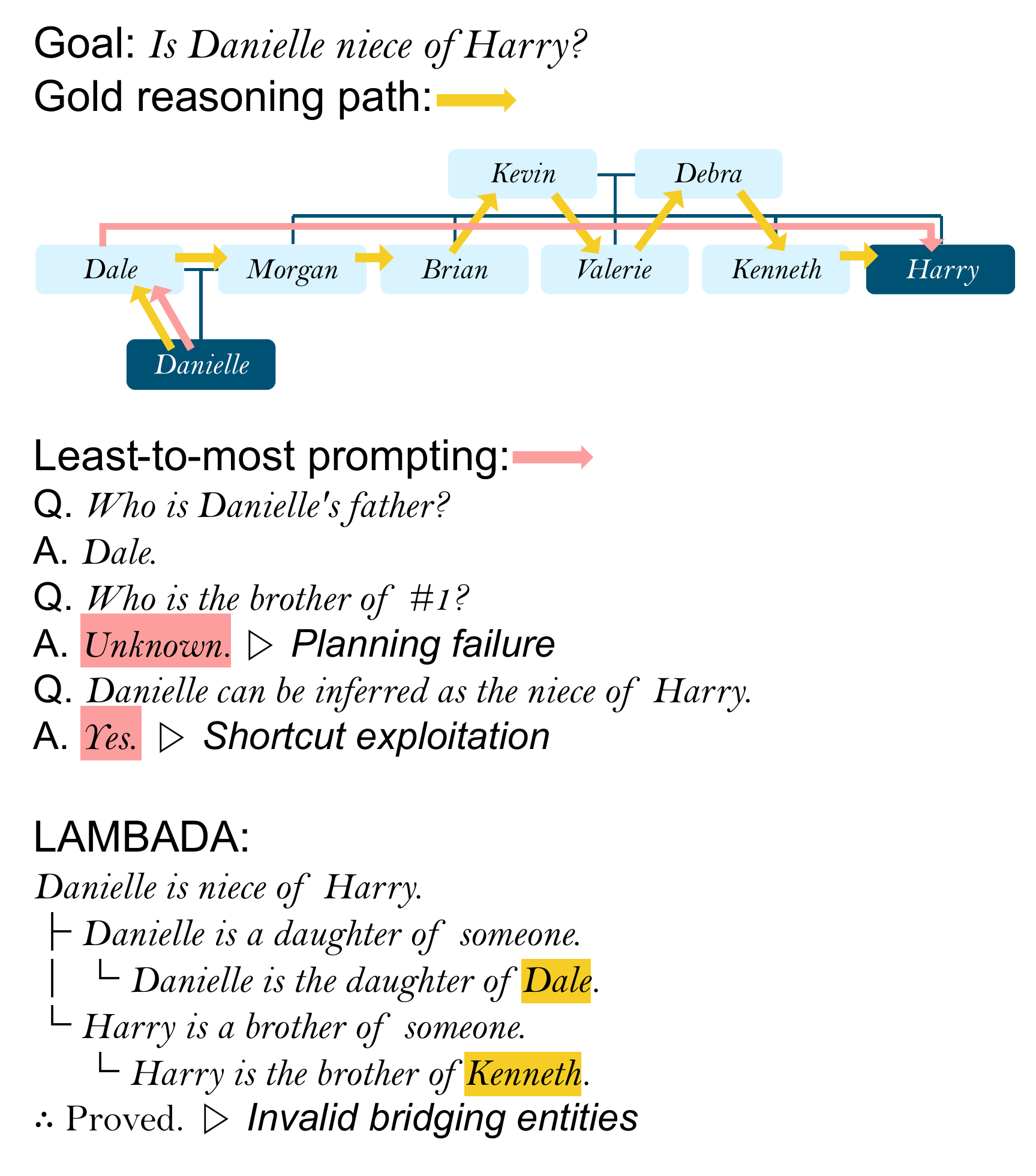}
    \caption{Example from CLUTRR. The proof is correct if it shows a chain of bridging entities, possibly omitting some. Least-to-most exploits shortcut, as it mispredicted the reasoning path but answered the final question correctly. LAMBADA cannot resolve the coreference between bridging entities, leading to disconnected proof.}
    \label{fig:l2m-shortcut}
\end{figure}

For Least-to-most, the low proof accuracy can be attributed to \textit{shortcuts}, where it fails to find an accurate decomposition but somehow reaches the correct answer. Figure \ref{fig:l2m-shortcut} illustrates the case where Least-to-most produces incorrect explanations.

In the case of LAMBADA, it cannot find the correct reasoning path if more than two bridging entities are involved in the proof (Figure \ref{fig:lambada}). LAMBADA's proof can only be accurate when there is zero or one bridging entity in the gold path, which is a \textit{coincidence} rather than a success.


\subsection{Efficiency}
\label{sec:efficiency}

To compare the efficiency of the compared methods, we report the token usage, API cost, and execution time for completing 300 examples in ProofWriter following \citet{kazemi-etal-2023-lambada}.

\begin{table}[t]
    \small
    \centering
    \begin{tabular}{l|c|c|c}
          & Tokens & Cost(\$) & Time(h) \\
          \hline \hline
         CoT           & 202,420 & 8.02 & 0.62 \\
         \hline
         Least-to-most & 1,485,989 & 47.14 & 1.18 \\
         LAMBADA       & 6,625,623 & 221.72 & 23.96 \\
         \ours         & \textbf{880,106} & \textbf{27.22} & \textbf{1.15} \\
    \end{tabular}
    \caption{Token/cost/time consumption (lower the better) for 300 examples in ProofWriter benchmark in GPT-4 Turbo. Regarding the cost, the OpenAI API used in this study charges \$0.03 per 1,000 input tokens and \$0.05 per 1,000 output tokens.}
    \label{tab:efficiency}
\end{table}


The results are presented in Table \ref{tab:efficiency}.  \ours \ achieves ~9x token/cost efficiency and ~22x speed compared to LAMBADA. While LAMBADA uses an LLM to perform decomposition and unification checks, these processes run symbolically in \ours, significantly reducing LLM inference cost.

Despite performing a \textit{complete} search when Least-to-most performs decomposition only once, \ours \ is even more efficient than Least-to-most prompting in ProofWriter. Although Least-to-most prompting can be optimized by dynamically appending the questions to intermediate sequences during the inference, currently available commercial LLM APIs do not support such functionality.

\section{Analysis}





\subsection{Solver ablation}
\label{sec:ablation-solver}

In previous sections, we show that Least-to-most's lack of backtracking reduces proof accuracy, and LAMBADA's lack of binding propagation restricts relational and arithmetic reasoning ability. However, the implementation details of \ours\ and the baselines are significantly different; \textit{e.g.} \ours\ uses logic programs as intermediate representations for reasoning. Therefore, we conduct an ablation study on \ours\ to refine the empirical effects of binding propagation and backtracking.

In this section, we directly manipulate the solver algorithm, while the LLM portion (single-step statement generation) remains as it is. In the \texttt{-Backtrack} setting, the symbolic solver will apply only one decomposition and binding even if there are multiple possible ways, as in Figure \ref{fig:baselines}(a). In the \texttt{-BindingProp} setting, the bindings obtained from previous subgoals are not propagated to subsequent ones, as in Figure \ref{fig:baselines}(b).

\begin{table}[ht]
    \small
    \centering
    \begin{tabular}{l|c|c|c|c}
          & PW & BE & CLUTRR & GSM8k \\
          \hline \hline
         \ours & 79.8 & 94.4 & 84.3 & 63.8 \\
         \hline
         \texttt{-Backtrack} & 76.3 & 82.9 & 69.8 & 62.0 \\
         (Least-to-most) & 71.5 & 83.4 & 81.5 & 60.6 \\
         \hline
         \texttt{-BindingProp} & 80.5 & 92.2 & 68.3 & 0.0 \\
         (LAMBADA) & 69.7 & 83.4 & 73.8 & 0.0
    \end{tabular}
    \caption{Answer accuracy (\%) of \texttt{-Backtrack} and \texttt{-BindingProp} for four benchmarks, experimented with GPT-4 Turbo. PW and BE stand for ProofWriter and Bird-Electricity, respectively. Results from Least-to-most and LAMBADA are also presented for reference.}
    \label{tab:ablation-solver}
\end{table}

The results are presented in Table \ref{tab:ablation-solver}. \texttt{-Backtrack} setting achieves significantly degraded performance in Birds-Electricity and CLUTRR by more than 10\%p, indicating that traversing multiple reasoning paths is crucial in these benchmarks. Compared to Least-to-most, \texttt{-Backtrack} performs better in ProofWriter but worse in CLUTRR. While Least-to-most exhibits low proof accuracy in both datasets (Figure \ref{fig:proofaccuracy}), Least-to-most tends to find a shortcut in CLUTRR that mitigates the effect of incorrect decomposition.

Analogous to LAMBADA, \texttt{-BindingProp} cannot answer GSM8k by design, as there is no way to pass the calculated results to the root goal. The \texttt{-BindingProp} outperforming LAMBADA in deductive benchmarks can again be attributed to negation handling.

\subsection{Single-step statement generation ablation}

While the SLD Resolution solver plays a key role in \ours, the implementation of single-step statement generation (LLM-based component) also affects \ours's performance. We perform ablation studies on the LLM-based modules and few-shot prompting methods. Due to space limits, the results are presented in Appendix \ref{sec:error}.

\section{Related works}

\subsection{Backward chaining in Natural Language Reasoning}

Backward chaining has not been explored much in the era of LLMs. At the time of writing, the only work that explicitly claims to be an LLM-based backward chaining system is LAMBADA \citep{kazemi-etal-2023-lambada}.

Alternatively, some backward chaining works use relatively small models directly fine-tuned with in-domain data \citep{tafjord-etal-2022-entailer, bostrom-etal-2022-natural, hong-etal-2022-metgen}. These methods train individual modules for rule generation and step verification, achieving strong results in its target domain but on behalf of the costly construction of in-domain training data.

Furthermore, as previously described in Section \ref{sec:baselines}, approaches based on task decomposition \citep{DBLP:conf/iclr/ZhouSHWS0SCBLC23, khot2023decomposed, radhakrishnan2023question} can be viewed as a type of backward chaining \citep{huang-chang-2023-towards}. Nonetheless, these methods tend to demonstrate relatively low proof accuracy due to decomposition failure \citep[Section \ref{sec:proof-accuracy} of this work]{radhakrishnan2023question}, while \ours \ is capable of providing a fully structured proof with high precision.

\subsection{LLM and Logic programming}

Integrating logic programming and LLMs for reasoning is a recently emerging topic \citep[\textit{inter alia.}]{pan-etal-2023-logic, yang2023neuro, olausson-etal-2023-linc}, triggered by the improvement in reasoning and code generation ability of LLMs. The majority of these works implement a similar two-stage approach: (1) convert the natural language reasoning task into a logic program, and (2) run an external solver to prove the query.


\ours \ differs from these methods as the solver is integrated into the loop instead of operating in separate stages. It is reported that LLMs often choose incompatible representations for the same concept or fail to discover information that does not surface in the premises \citep{olausson-etal-2023-linc}, as they generate the code without any hierarchical cues about how statements are structured. These issues can be potentially mitigated by the backward chaining of \ours, as it ensures that all subgoals are addressed at least once by backtracking and that the generated statement unifies with the query by the Symbolic Verification module.

\section{Conclusion}

While backward chaining is a promising direction for structured natural language reasoning, current LLM-based approaches like Least-to-most and LAMBADA are only incomplete reproductions of backward chaining as they leave out backtracking and binding propagation. To this extent, we build \ours\ directly from the SLD Resolution algorithm. In \ours, a symbolic solver controls the proof, while an LLM searches and translates relevant natural language statements into symbolic representations.


\ours\ outperforms backward chaining baselines in diverse reasoning tasks including deductive, relational, and arithmetic reasoning. Not only does it reach the correct answer more frequently, but also demonstrates improved proof accuracy and efficiency than baselines. From both theoretical and empirical perspectives, we believe that \ours \ significantly extends the horizon of LLM-based backward chaining.

\section{Limitations}

While \ours \ significantly improves the completeness, performance, and efficiency of LLM-based backward chaining, it still holds limitations inherited from backward chaining, symbolic reasoning, and LLMs.

Even though backward chaining proof always \textit{terminates}, a naively implemented backward chaining system might still require substantial computation in fact-intensive tasks such as knowledge base question answering (KBQA) \citep{yih-etal-2016-value, gu2021beyond}. This might be mitigated by hybrid forward and backward chaining \citep{hong-etal-2022-metgen} or by using sophisticated planning algorithms for symbolic solvers \citep{DBLP:conf/ccis/LuLG12, yang2023neuro}. We leave this direction as future work.

Furthermore, some reasoning problems may not be able to be effectively formulated in logic programming notations as in this study. Most notably, solving high-order logic problems generally requires \textit{meta-predicates} that reason over the database, such as \texttt{call/N} in Prolog \citep{DBLP:journals/jlp/ChenKW93}, which cannot be handled using the first-order SLD Resolution algorithm of \ours. Besides high-order logic, some reasoning tasks \citep[\textit{e.g.}][]{dalvi-etal-2021-explaining, zellers2019hellaswag} require reasoning with complex linguistic expressions and highly pragmatic assumptions, which might not be effectively expressed using logic programming.

Finally, LLMs often produce counterfactual and inconsistent information, and can potentially cause risk when used in domains where high precision and factuality are required. While \ours \ reduces errors by leveraging the symbolic solver and applying a modular approach, the single-step statement generation based on LLM is still subjective to producing false reasoning steps that might lead to the wrong conclusion.

\section{Acknowledgements}

We would like to express our sincere gratitude to Hyunjun Kim for his invaluable advice in the initial stages of the research. We also thank Julia Hockenmaier and Takyoung Kim for their constructive suggestions and encouragement, which greatly contributed to the completion of this work.

\vspace{3cm} 

\bibliography{anthology,custom}
\bibliographystyle{acl_natbib}

\clearpage

\appendix

\section{Formal definition of \ours}
\label{sec:appendix-solver}

In this section, we provide an algorithmic description of \ours. \ours \ can be viewed as an extension of the SLD Resolution (Selective Linear Definite Resolution) algorithm \citep{kowalski1974predicate}, which is the staple of modern logic programming languages like SWI-Prolog \citep{wielemaker:2011:tplp}. A pseudo-code for \ours \ is presented in Algorithm \ref{alg:topdownsolver}. The notations used throughout this section are presented in Table \ref{tab:notations}.

\begin{table}[ht]
    \small
    \centering
    \begin{tabularx}{\columnwidth}{XXX}
         \multicolumn{1}{c}{Notation} & \multicolumn{1}{c}{Definition} \\
         \hline \hline
         \multicolumn{1}{c}{$h, p, q$} & Terms \\
         \multicolumn{1}{c}{$\mathbb{T}$} & Set of all terms \\
         \multicolumn{1}{c}{$B(p)$} & A proved binding for term $p$. \\
         \multicolumn{1}{c}{$\mathcal{B}(p)$} & List of all proved bindings for term $p$. \\
         \multicolumn{1}{c}{$\mathbb{B}$} & Set of all bindings. \\
         \multicolumn{1}{c}{\textbf{r}} & Rule \\
         \multicolumn{1}{c}{$\textrm{\textbf{r}}.head$} & Rule head (term) \\
         \multicolumn{1}{c}{$\textrm{\textbf{r}}.subgoal$} & Rule subgoals (list of terms) \\
         \multicolumn{1}{c}{$\textsc{NL}$} & Natural language description of rules\\
    \end{tabularx}
    \caption{Notations used in Appendix \ref{sec:appendix-solver}. Note that facts are special instances of rules where $|\textrm{\textbf{r}}.subgoal| = 0$.}
    \label{tab:notations}
\end{table} 

Before proceeding to the algorithm, we introduce three procedures about unification and binding, namely $\textsc{Unify}: \mathbb{T} \times \mathbb{T} \rightarrow \{0,1\}$, $\textsc{Binding}: \mathbb{T} \times \mathbb{T} \rightarrow \mathbb{B}$, and $\textsc{Bind}: \mathbb{T} \times \mathbb{B} \rightarrow \mathbb{T}$. As described in Section \ref{sec:backgroundsolver}, two terms are said to unify if there is a valid binding that makes the terms identical. \textsc{Unify} returns a boolean value indicating whether the two terms unify or not. \textsc{Binding} returns the binding of two terms if they unify. \textsc{Bind} takes a term (possibly containing variables) and a binding as its argument, and returns the bound term after substituting the variables from the term to the corresponding values. By definition, for any two terms $p$ and $q$ that satisfy $\textsc{Unify}(p, q)$, $\textsc{Bind}(p, \textsc{Binding}(p, q)) = \textsc{Bind}(q, \textsc{Binding}(p, q))$ should always hold.

\textsc{Solve} is the main procedure of \ours. It receives a goal term $q$ as a parameter and refers to the global database $\mathcal{D}$ to compute $\mathcal{B}(q)$, the list of all provable bindings for $q$. If $\mathcal{B}(q)$ is not empty, it implies that $q$ can be proved on $\mathcal{D}$. Otherwise, the goal cannot be proved.


The main loop, which performs a combinatorial search for every possible binding, is shown in Lines 5-19. First, rules that unify with the goal are selected from the database (Line 4, \textbf{Search}). The initial binding $B_0$ is the binding between the rule head and the goal. We iterate through the subgoals (Line 7, \textbf{Decompose}) to perform a complete search. For each subgoal $p_t$, we bind the subgoal using the previous binding $B(p_{t-1})_i$ (Line 10, \textbf{Binding propagation}). The partially bound subgoal $p_{t, i}$ is proved by recursively calling \textsc{Solve}, which returns a list of bindings $B(p_{t, i})$ for $p_{t, i}$ (Line 11). The binding $B(p_{t, i})$ is updated to the new bindings $B(p_{t, i})_j$ (Line 15), which will be propagated to the next subgoal $p_{t+1}$.

After testing all subgoals, if $\mathcal{B}(p_T)$ is non-empty, we can conclude that $q$ is proved with respect to the binding. In constant, if any subgoal $p_t$ is not provable, $\mathcal{B}(p_T)$ will eventually be empty. However, as we are iterating through all unifying rules (Line 5, \textbf{Backtracking}), \textsc{Solve} will proceed to other possible decompositions.

Single-step statement generation, the novel mechanism of \ours, is shown in Lines 23-25. When the binding for goal $q$ and all subgoals $p_i$ is found, proof has succeeded and \textsc{Solve} returns the binding. However, when the proof has failed, the single-step statement generation (\textsc{SingleStepStmtGen}) process described in Section \ref{sec:ours-method} is called, returning a new statement $\textrm{\textbf{r}}_{new}$ from the natural language description $NL$ and the goal $q$. If the procedure succeeds, $\textrm{\textbf{r}}_{new}$ is added to $\mathcal{D}$, and the solver re-attempts to solve $q$ with the updated database.


For brevity, here we do not further describe extensions and optimizations, namely Negation-as-failure \citep{apt1992new}, arithmetic and comparison operators, odd loop on negation (OLON) \citep{DBLP:journals/corr/abs-1709-00501}, goal tabling, and proof tree generation. Full implementation of \ours \ can be found in this \href{https://github.com/lbox-kr/symba}{repository}.

\begin{algorithm*}
\caption{Algorithm of \ours}\label{alg:topdownsolver}
\begin{algorithmic}[1]
\State \textbf{global } $\mathcal{D} \gets$ \textit{empty set}, $\textsc{NL} \gets$ \textit{natural language description}
\Procedure{Solve}{$q$} \Comment{Input: goal term, Returns: list of bindings}
    \State $\mathcal{B}(q) \gets$ \textit{empty list}
    \State $\mathcal{R} \gets \{\textrm{\textbf{r}}\in\mathcal{D} \ | \ \textsc{Unify}(\textrm{\textbf{r}}.head, q)\}$ \Comment{\textbf{Search}: find unifying rules from database}
    \For{$\textrm{\textbf{r}}\in\mathcal{R}$} \Comment{\textbf{Backtracking}: If a rule fails, try another}
        \State $\mathcal{B}(\textrm{\textbf{r}}.head) \gets [\textsc{Binding}(\textrm{\textbf{r}}.head, q)]$
        \For{$p_t\in \textrm{\textbf{r}}.subgoal = [p_1, ..., p_{T}]$} \Comment{\textbf{Decompose}: Iterate through subgoals}
        \State $\mathcal{B}(p_{t}) \gets$ \textit{empty list}
            \For{$B(p_{t-1})_{i}\in \mathcal{B}(p_{t-1})$}
                \State $p_{t,i} \gets \textsc{Bind}(p_t, B(p_{t-1})_{i})$ \Comment{\textbf{Binding Propagation}: Bind $p_{t}$ with previous bindings}
                \State $\mathcal{B}(p_{t})_{i} \gets \textsc{Solve}(p_{t,i})$
                \For{$B(p_{t,i})_{j} \in \mathcal{B}(p_{t,i})$}
                    \State $B(p_{t,i})_{j} \gets B(p_{t,i})_{j} \cup B(p_{t-1})_i$
                \EndFor
                \State Extend $B(p_{t})_i$ to $\mathcal{B}(p_{t})$ \Comment{Accumulate bindings for propagation}
            \EndFor
        \EndFor
        \State Extend $B{p_T}$ to $\mathcal{B}(q)$
    \EndFor
    \If{$B{p_T}$ is not empty} \Comment{Proof success}
        \State \Return $\mathcal{B}(q)$
    \Else  \Comment{Proof fail}
        \State $\textrm{\textbf{r}}_{new} \gets \textsc{SingleStepStmtGen}(\textsc{NL}, q)$
        \State Add $\textrm{\textbf{r}}_{new}$ to $\mathcal{D}$
        \State \Return{$\textsc{Solve}(q)$}
    \EndIf
\EndProcedure
\end{algorithmic}
\end{algorithm*}
\section{Dataset details}
\label{sec:appendix_dataset}

\begin{table*}[ht]
    \small
    \centering
    \begin{tabular}{l|c|c|c|c|c|c }
         \multicolumn{1}{c|}{Dataset} & Type & \# examples & Avg. steps & Avg. sents & N-shot \\
         \hline \hline
         ProofWriter \citep{tafjord-etal-2021-proofwriter} & Deductive & 300 & 4.52 & 19.12 & 3\\
         Birds-Electricity (\textit{Ibid.}) & Deductive & 300 & 2.08 & 13.77 & 3\\
         ParaRules \citep{ijcai2020p537} & Deductive & 300 & 4.37 & 10.56 & 3\\
         PrOntoQA \citep{DBLP:conf/iclr/Saparov023} & Deductive & 100 & 4.00 & 21.84 & 3\\
         CLUTRR \citep{sinha-etal-2019-clutrr} & Relational & 100 & 4.86 & 5.20 & 3\\
         MAWPS \citep{koncel-kedziorski-etal-2016-mawps} & Arithmetic & 300 & 3.06 & 3.20 & 5 \\
         GSM8k \citep{cobbe2021gsm8k} & Arithmetic & 270 & 9.22 & 4.87 & 5
    \end{tabular}
    \caption{Statistics of each test set. \textit{\# examples} indicates the number of sampled examples from the original test set, due to budget constraints. \textit{Avg. steps} denotes the average number of statements (facts and rules) required to prove the goal, and \textit{Avg. sents} is the average number of sentences that each context contains. \textit{N-shot} denotes the number of few-shot examples to prompt LLMs in this study.}
    \label{tab:dataset}
\end{table*}



This section describes the sampling, preprocessing, and evaluation of benchmarks. Table \ref{tab:dataset} presents brief information and statistics about the seven benchmarks used in this paper.

All datasets used in this study allow free use, modification, and redistribution for non-commercial applications.


\subsection{ProofWriter family}

\hspace{\parindent}\textbf{Test split sampling} From the ProofWriter family, we sample the evaluation set from the test split of the closed-world assumption subset (CWA). Specifically, for ProofWriter, we use the dep5 subset, which has a deepest maximum reasoning depth of 5. Since a single context includes multiple questions, we first sample 300 contexts and randomly sample a question from it. As a result, we obtain 300 (context, question) tuples for each dataset.

\textbf{In-context demonstrations} We randomly sample 3 examples from ProofWriter-dep3 and -dep2 data that contain shorter contexts to test the length generalization ability of each method. For CoT prompting and Least-to-most prompting, we provide the pre-order traversal of the golden proof tree provided for each instance, with stopwords like \textit{since} and \textit{so} that are known to enhance the performance in CoT prompting \citep{kazemi-etal-2023-lambada}. For LAMBADA, we use the prompt format provided in the original paper, which is populated with the sampled in-context examples.

\textbf{Logic program} We consistently apply $\texttt{verb}(\texttt{subject}, \texttt{object})$ format to all datasets. For instance, \textit{Bald eagle does not eat the mouse.} translates to $\texttt{not eats}(\texttt{bald\_eagle}, \texttt{mouse})$. Note that we apply the same format for adjective facts. For example, the corresponding symbolic form for \textit{Alan is young.} is $\texttt{is}(\texttt{alan}, \texttt{young})$, opposed to another commonly used form $\texttt{young}(\texttt{alan})$ or $\texttt{young}(\texttt{alan}, \texttt{true})$ \citep{olausson-etal-2023-linc, pan-etal-2023-logic}.

As a common practice for measuring the reasoning ability in out-of-distribution data (Birds-Electricity, ParaRules) using in-domain data (ProofWriter) \citep{tafjord-etal-2021-proofwriter}, we use the prompts and examples sampled from ProofWriter train split for the other two benchmarks.

\textbf{Evaluation} We use the true/false labels provided with the original dataset without modification.

\subsection{PrOntoQA}

\hspace{\parindent}\textbf{Test split sampling} We sample the test set using the original script from \citet{DBLP:conf/iclr/Saparov023}, using fictional entity names (\textit{e.g.} \textit{Every yumpus is a jompus.}). However, due to an unresolved issue of the script, the script only allows to generate a reasoning chain of a maximum of four steps. 

\textbf{In-context demonstrations} Similar to the ProofWriter family, we use few-shot demonstrations with 8 premises, which is significantly lower than average (21.84 premises).

We use identical logic program formats and evaluation criteria for PrOntoQA with other ProofWriter variants.

\subsection{CLUTRR}

\hspace{\parindent}\textbf{Test split sampling} We randomly sample 100 examples from the test split of CLUTRR v1. To generate false labels, we sample half of the examples and alter the relation label of the gold triplet to a random one.

\textbf{In-context demonstrations} We randomly sample 3 stories from the train split that only contains 2-3 relations to test the length generalization ability of each method. For CoT, we provide a golden chain of kinship relations that connect the two queried entities. For Least-to-most prompting, each decomposed question contains information about an entity and a relation, asking for the bridging entity. (e.g. \textit{Who is the father of Andrea?})

\textbf{Rules} To minimize the effects of pretrained knowledge, we append 39 rules about family relationships to each story, \textit{e.g.} \textit{If A is B's son and B is C's son, A is C's grandson}.

\textbf{Logic program}
To prevent infinite recursion, we use separate predicate names for the base fact and inferred relations. For instance, '\textit{George is the father of Andrea.}' is translated as $\texttt{isRelationOf}(\texttt{george}, \texttt{father}, \texttt{andrea})$ if it is a fact directly from the context, or $\texttt{relation}(\texttt{george}, \texttt{father}, \texttt{andrea})$ if it is inferred by more than one bridging entities. Note that the predicate name for the latter casts no effect on the single-step statement generation's performance as it is only used for the symbolic solver and not the LLM.

\textbf{Evaluation} Each model is instructed to predict if the given relation holds between the two entities. Half of these tuples have correct relation labels, and the other half have randomized labels that preserve the gender of the correct answer.

\subsection{MAWPS}

\hspace{\parindent}\textbf{Test split sampling} We use the first 300 examples from the original test split. 

\textbf{In-context demonstrations} Five few-shot examples are randomly sampled from the train split. We manually create annotations as the benchmark does not include a reasoning chain.

\textbf{Logic program} We denote the meaning of each numeric value with predicates of arity 1, as in $\texttt{number\_of\_oranges}(\_)$ or $\texttt{fraction\_of\_trombone\_section}(\_)$. We use $\texttt{answer}(X)$ to express the final answer in all examples and evaluate if the variable $X$ is successfully bound to the right numeric value (\textit{e.g.} $\texttt{answer}(5)$).\footnote{While previous approaches in logic programming-integrated LLMs use an additional step to specify which predicate corresponds to the final answer \citep{pan-etal-2023-logic}, we do not introduce this mechanism for uniformness between tasks.} Facts denote the base value mentioned in the text (\textit{e.g.} $\texttt{number\_of\_yellow\_flowers}(10)$), and rules express the arithmetic relations between each value (\textit{e.g.} \texttt{fraction\_of\_trumpet\_section}(X) :- \texttt{fraction\_of\_trombone\_section}(A), $X = A * 4.$).

\textbf{Evaluation} We use the numeric answer provided with the original dataset. If the answer is not a numeric string (e.g. $\texttt{25,000}$ or $\texttt{42 pages}$), they are considered incorrect. 

\subsection{GSM8k}

\hspace{\parindent}\textbf{Test split sampling} We use the test split used in \citet{yang2023neuro}, which contains 270 examples and is a subset of the original test split from \citet{cobbe2021gsm8k}. We calculate the number of reasoning steps presented in Table \ref{tab:dataset} based on the semi-structured solutions included in the dataset.

\textbf{In-context demonstrations} We randomly sample 5 questions from the train split. For CoT prompting, we used the \texttt{answer} column from the original dataset and removed the external call snippets (equations that are wrapped in double angle brackets <<...>>). For Least-to-most prompting, we reformulate the \texttt{answer} column from the `Socratic' version of the dataset that formulates the reasoning chain as consecutive sequence of questions and answers.

We use identical logic program formats and evaluation criteria for GSM8k with MAWPS.

\section{Analysis on Single-step statement generation}
\label{sec:error}

\subsection{Negative few-shot examples}
\label{sec:negative-examples}

\begin{figure}[!tb]
    \centering
    \includegraphics[width=0.8\columnwidth]{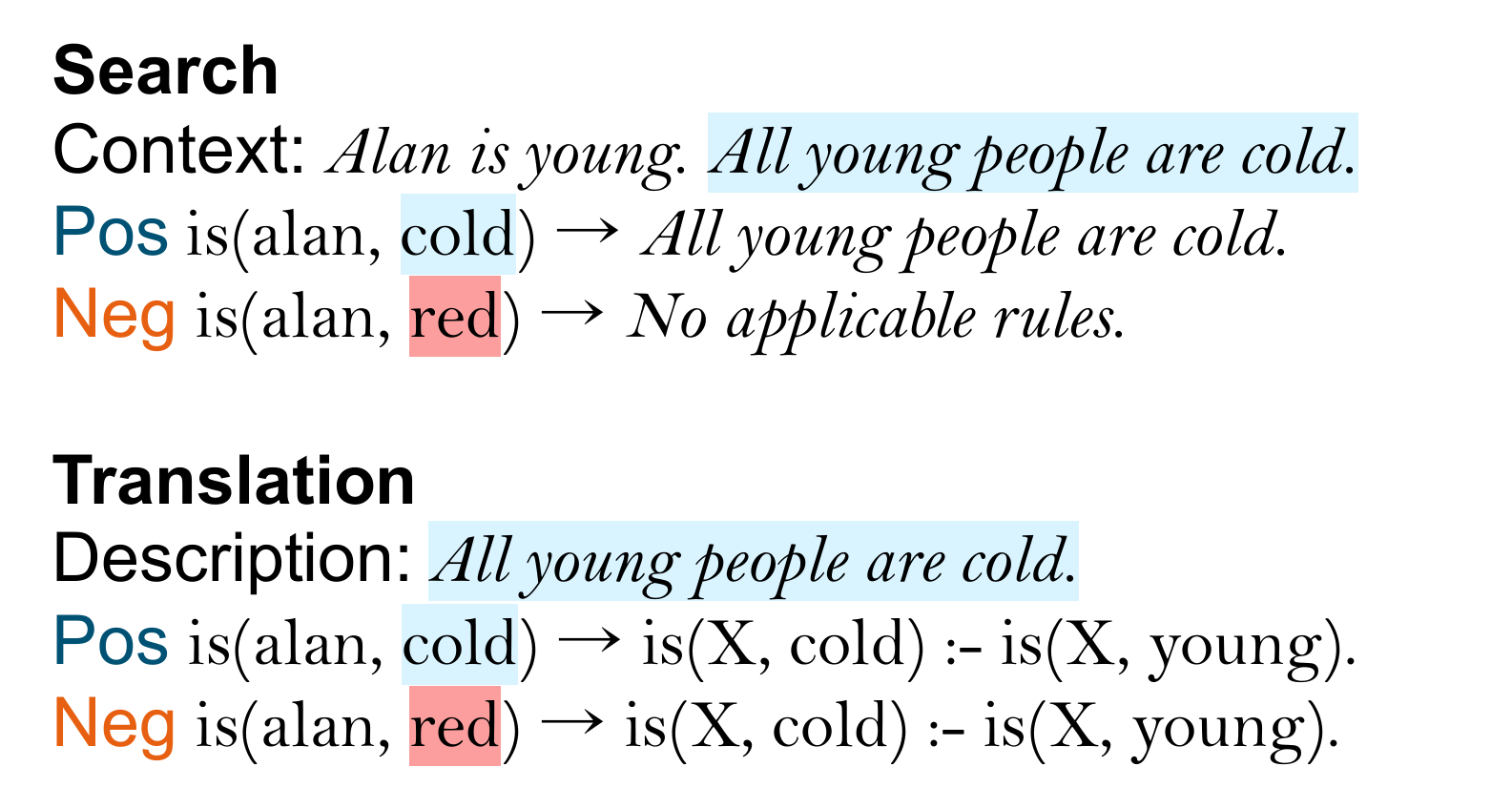}
    \caption{Examples of Positive/Negative examples included in the prompts for the Search/Translation module of \ours.}
    \label{fig:prompt-examples}
\end{figure}

In the preliminary experiments, we observe that LLMs often generate hallucinated outputs that follow the symbolic goal but are not stated in the natural language problem. To mitigate the issue, we combine the Positive and Negative examples to reduce hallucination in the Search/Translation modules (Figure \ref{fig:prompt-examples}). Negative examples are generated by creating a mismatch between the symbolic and natural language inputs so the LLMs can follow the content of the natural language.

\subsection{Ablation study}
\label{sec:ablation-fewshot}

As an ablation study, we selectively manipulate the modules or in-context demonstrations and examine the performance of four tasks.

\textbf{Modules} To analyze the contribution of each module, we selectively remove some and compare the performance. In the \texttt{-Search} setting, we remove Fact/Rule Search by merging it to Fact/Rule Translation, so that the symbolic statement is directly generated from the context and the query without intermediate textual representations. In the \texttt{-Unify} setting, we disable the Symbolic Validation module by not checking if the generated statement unifies to the query.

\textbf{Negative in-context examples} We also test the effects of the Negative in-context examples illustrated in Figure \ref{fig:prompt-examples}. In the \texttt{-SearchNeg} setting, we remove Negative examples from the Search module, while in \texttt{-TransNeg} we remove Negative examples from the Translation module.

\begin{table}[ht]
    \small
    \centering
    \begin{tabular}{l|c|c|c|c}
          & PW & BE & CLUTRR & GSM8k \\
          \hline \hline
         \ours & 79.8 & 94.4 & 84.3 & 63.8 \\
         \hline
         \texttt{-Search} & -22.7 & -5.2 & +2.4 & +3.0 \\
         \texttt{-Unify} & -6.9 & -1.6 & -8.7 & -0.1 \\
         \hline
         \texttt{-SearchNeg} & -8.8 & -29.8 & +2.7 & +4.1 \\
         \texttt{-TransNeg} & -2.4 & -12.0 & -13.8 & +1.5 \\
    \end{tabular}
    \caption{Ablation results on four benchmarks using GPT-4 Turbo. All ablation results are 4-run.}
    \label{tab:ablation}
\end{table}

\begin{figure}[!tb]
    \centering
    \includegraphics[width=\columnwidth]{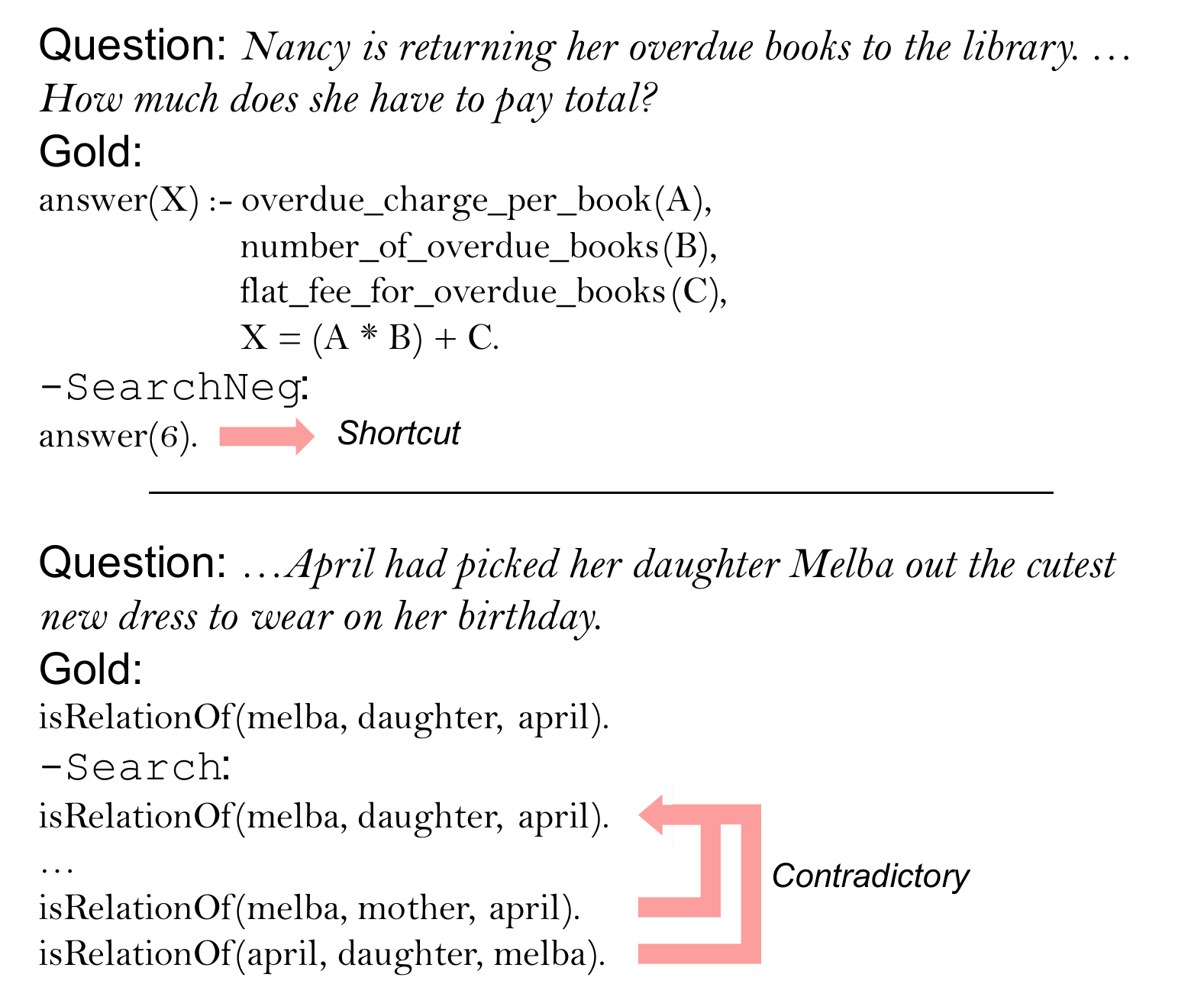}
    \caption{Examples of erroneous logic program statements, sampled from \texttt{-SearchNeg} in GSM8k and \texttt{-Search} in CLUTRR. Ablated versions often fail to produce a faithful reasoning path where \ours \ generates a correct proof (denoted as Gold).}
    \label{fig:ablation-examples}
\end{figure}

As presented in Table \ref{tab:ablation}, each ablation leads to a significant performance drop in specific benchmarks, especially in ProofWriter variants, indicating that modules and negative in-context examples are necessary components of \ours. While some ablation settings achieve similar or even better performance in CLUTRR and GSM8k, we observe common issues related to the proof accuracy in these settings (Figure \ref{fig:ablation-examples}).

\section{Error analysis}

We manually classify the LLM module errors observed from \ours \ into three categories: Search-Hallucination, Search-Miss, and Translation. Definitions of the error types are shown in Table \ref{tab:errortype}.

\begin{table}[ht]
    \small
    \centering
    \begin{tabularx}{\columnwidth}{XX}
         \multicolumn{1}{c}{Error Type} & \multicolumn{1}{c}{Definition} \\
         \hline \hline
         \multicolumn{1}{c}{Search-Hallucination} & The generated description is not in the context, or unrelated to the query. \\
         \hline
         \multicolumn{1}{c}{Search-Miss} & A relevant description stated in the context was not retrieved. \\
         \hline
         \multicolumn{1}{c}{Translation} & Symbolic statement is unfaithfully translated from the description (\textit{i.e.} syntax error, misleading symbol names).
    \end{tabularx}
    \caption{Description of three error classes observed from \ours. If multiple errors occur simultaneously in one example, we select the error that appears first.}
    \label{tab:errortype}
\end{table}

\begin{figure}[!tb]
    \centering
    \includegraphics[width=\columnwidth]{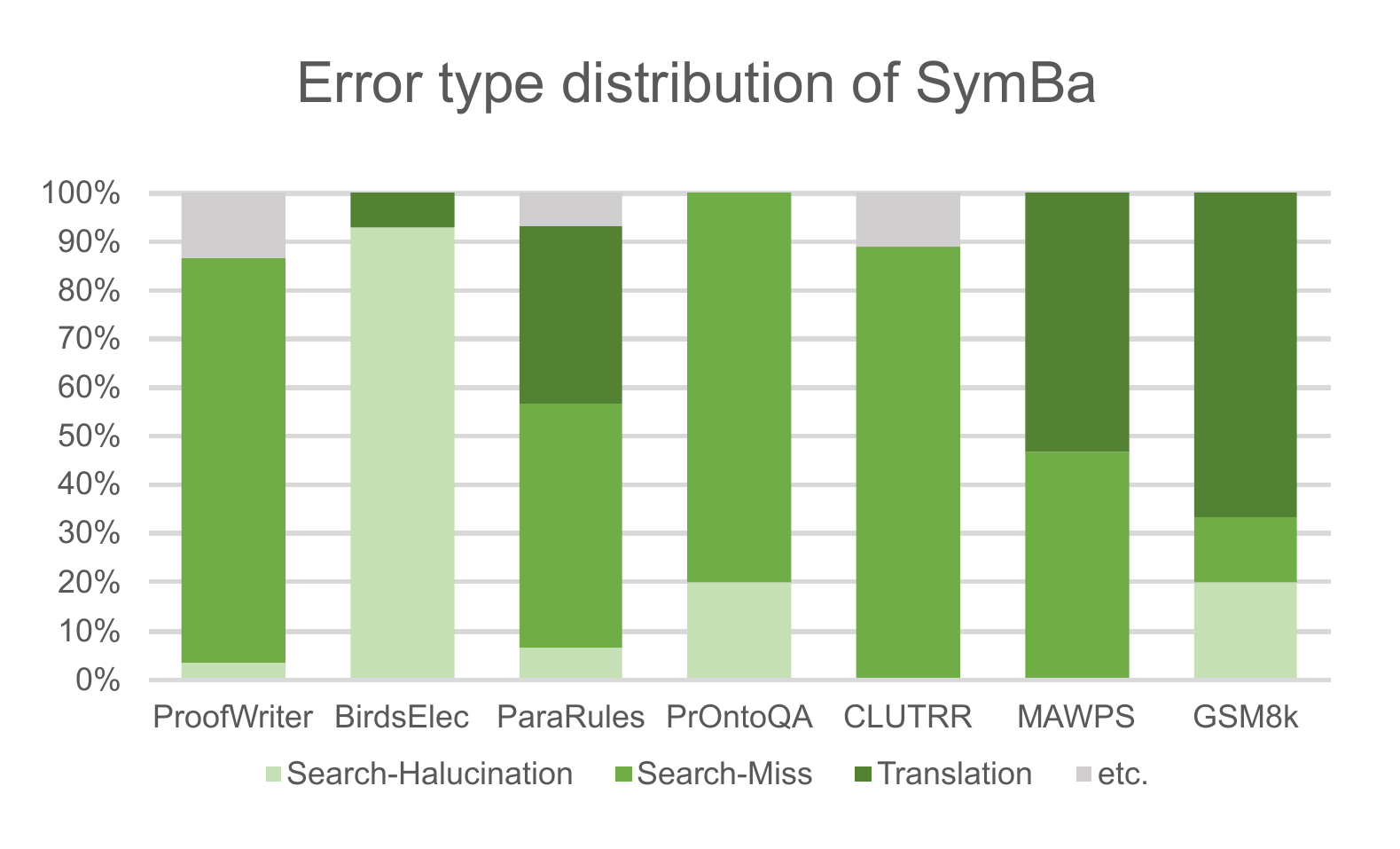}
    \caption{Error analysis results for \ours. 30 incorrect proofs are sampled and manually classified according to Table \ref{tab:errortype}.}
    \label{fig:errorheatmap}
\end{figure}

As presented in Figure \ref{fig:errorheatmap}, the distribution of errors highly varies along the datasets. It implies that each benchmark poses unique challenges depending on numerous factors, such as reasoning type and lexical diversity.

Among the benchmarks, we focus on ProofWriter and Birds-Electricity, which are syntactically near-identical yet display completely different error distributions. While rules in ProofWriter often contain variables ('\textit{If \textbf{someone} is red then they are round}'), 99.6\% of the rules from Birds-Electricity are bound ('\textit{If wire is metal then wire conducts electricity}'). From this observation, we hypothesize that the higher ratio of unbound rules leads to elevated Search-miss errors.

\begin{figure}[!htb]
    \centering
    \includegraphics[width=0.95\columnwidth]{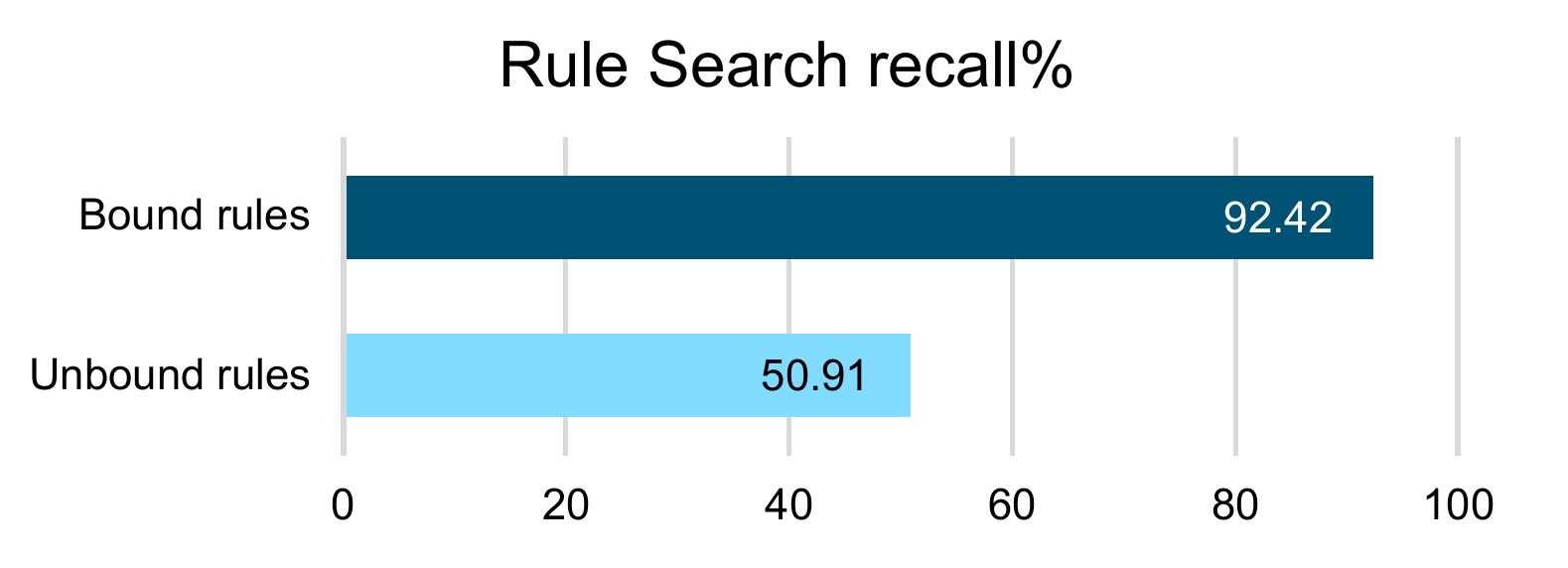}
    \caption{Recall of the Rule Search module in bound and unbound ProofWriter rules.}
    \label{fig:errorquant}
\end{figure}

We compare the recall of the Rule Search module in isolation, based on whether the target rule is bound or not (Figure \ref{fig:errorquant}). Rule Search achieves a recall of approximately 51\% when the target rule is not bound, which is significantly lower than that of bound rules ($\sim$92\%). It shows that the boundness of the rules seriously affects Search-Miss errors, possibly due to the low lexical overlap of unbound rules compared to bound rules \citep{shinoda-etal-2021-question, ijcai2020p501}.

\begin{table*}[t]
    \small
    \centering
    \begin{tabular}{l|l|c|c|c|c|c|c|c}
        \multirow{2}{*}{Model} & \multirow{2}{*}{Method} & \multicolumn{7}{c}{Performance}\\
        \cline{3-9}
          &  & ProofWriter & BirdsElec & ParaRules & PrOntoQA & CLUTRR & MAWPS & GSM8k \\
          \hline \hline
          
         \multirow{5}{*}{GPT-4}
         & Standard & 63.2\stdev{0.43} & 77.8\stdev{1.17} & 61.3\stdev{1.10}& 83.0\stdev{0.82}& 72.0\stdev{4.00}& $^\dag$94.2\stdev{0.58}& 29.4\stdev{1.81}\\
         & CoT & 70.5\stdev{2.13}& 81.2\stdev{1.41}& 60.5\stdev{1.03}& 96.8\stdev{1.26}& $^\dag$84.5\stdev{1.29}& $^\dag$99.1\stdev{0.49}& $^\dag$94.2\stdev{1.00}\\ 
        \cline{2-9}
         & Least-to-most & 71.5\stdev{2.10}& 88.2\stdev{0.76}& 71.8\stdev{0.71}& 87.5\stdev{1.29}& 81.5\stdev{0.58}& 84.3\stdev{0.56}& 60.6\stdev{1.96}\\
         & LAMBADA & 69.7\stdev{1.18}& 83.4\stdev{1.20}& 59.7\stdev{1.30}& 96.0\stdev{1.41}& 73.8\stdev{1.50} & 0.0\stdev{0.00} & 0.0\stdev{0.00} \\
         & \ours & \textbf{79.8\stdev{1.06}}& \textbf{94.4\stdev{0.62}}& \textbf{79.2\stdev{1.12}}& \textbf{96.3\stdev{1.26}}& \textbf{84.3\stdev{2.06}}& \textbf{86.7\stdev{0.69}} & \textbf{63.8\stdev{0.74}}\\
        \hline \hline
        
         \multirow{5}{*}{Claude-3}
         & Standard & 61.3\stdev{0.00}& 66.0\stdev{0.00}& 61.3\stdev{0.00}& $^\dag$96.0\stdev{0.00}& 80.0\stdev{0.00}& $^\dag$96.3\stdev{0.00}& 17.0\stdev{0.00}\\
         & CoT & 67.0\stdev{2.00}& 73.3\stdev{0.00}& 57.3\stdev{0.00}& $^\dag$96.0\stdev{0.00}& 67.0\stdev{0.00}& 88.0\stdev{0.00}& $^\dag$92.2\stdev{0.00}\\
        \cline{2-9}
         & Least-to-most & 60.3\stdev{0.00} & 75.7\stdev{0.00} & 57.3\stdev{0.00} & 86.0\stdev{0.00} & 67.0\stdev{0.00} & \textbf{94.2\stdev{0.15}} & 59.3\stdev{0.00}\\
         & LAMBADA & 69.3\stdev{0.00}& 62.7\stdev{0.00} & 57.7\stdev{0.00} & 67.0\stdev{0.00} & 69.0\stdev{0.00} & 0.0\stdev{0.00} & 0.0\stdev{0.00} \\
         & \ours & \textbf{77.6\stdev{0.00}} & \textbf{77.3\stdev{0.00}} & \textbf{69.0\stdev{0.00}} & \textbf{91.0\stdev{0.00}} & \textbf{85.0\stdev{0.00}} & \textbf{94.1\stdev{0.15}} & \textbf{67.4\stdev{0.00}} \\
        \hline \hline
        
         \multirow{5}{*}{LLaMa-3}
         & Standard & 63.6\stdev{0.50}& 78.7\stdev{0.00}& 65.3\stdev{0.00}& $^\dag$99.0\stdev{0.00}& 75.0\stdev{0.00}& $^\dag$96.3\stdev{0.00}& 26.2\stdev{0.00}\\
         & CoT & 64.8\stdev{1.26}& 79.0\stdev{1.29}& 63.0\stdev{1.67} & 92.5\stdev{4.12}& 77.0\stdev{0.00} & $^\dag$95.0\stdev{0.00} & $^\dag$89.5\stdev{1.35}\\
        \cline{2-9}
         & Least-to-most & 61.4\stdev{0.34} & 71.0\stdev{0.00} & 66.7\stdev{0.00} & \textbf{95.0\stdev{0.00}} & 72.0\stdev{0.00} & \textbf{89.0\stdev{0.00}} & 61.5\stdev{0.00} \\
         & LAMBADA & 64.0\stdev{1.63}&82.3\stdev{0.00}& 62.1\stdev{1.10} & 90.8\stdev{0.50} & 73.3\stdev{0.50} & 0.0\stdev{0.00} & 0.0\stdev{0.00} \\
         & \ours & \textbf{70.4\stdev{1.26}} & \textbf{92.9\stdev{1.10}}& \textbf{71.7\stdev{0.00}}& 93.3\stdev{0.50}& \textbf{90.5\stdev{0.58}} & 87.9\stdev{0.70} & \textbf{67.0\stdev{0.00}}\\
    \end{tabular}
    \caption{Average accuracy (\%) and standard deviation on 4-runs per each benchmark and reasoning methods. Boldface font indicates that the score is significantly higher than other backward chaining methods, which is equivalent to the boldface in Table \ref{tab:main}. Daggers represent that non-structured methods (Standard, Chain-of-thought) achieve significantly higher score than the best structured backward chaining results. 95\% confidence applies to both notations. Note that the temperature was set to 0 for all runs, which results in zero standard deviation in some settings even when the seed is different.}
    \label{tab:full}
\end{table*}


\section{Complete results}
\label{sec:appendix_fullresult}

Table \ref{tab:full} presents the complete results of the main experiment (Section \ref{sec:task-performance}). We also report the performance of Standard prompting (generating the answer without any rationales) and Chain-of-thought prompting for comparison.

\end{document}